\documentclass[letterpaper, 10 pt]{IEEEtran}
\IEEEoverridecommandlockouts                              
                                                          
\IEEEoverridecommandlockouts 

\usepackage[noend, boxruled, linesnumbered] {algorithm2e}
\SetAlCapSkip{1em}
\makeatletter
\renewcommand{\algorithmcfname}{Alg.}
\renewcommand{\fnum@algocf}{\AlCapSty{\AlCapFnt\algorithmcfname\nobreakspace\thealgocf}}
\makeatother

\usepackage[noadjust,space]{cite}
\usepackage{svg}
\usepackage{amsmath,amsfonts}
\usepackage{algorithmic}
\usepackage{graphicx}
\usepackage{textcomp}
\usepackage{xcolor}
\usepackage{tabularx}
\usepackage{dsfont}
\usepackage[]{subcaption}
\usepackage{colortbl} 
\usepackage{hhline} 
\usepackage[colorlinks=true,allcolors=.,urlcolor=blue,]{hyperref}
\usepackage{multicol}
\usepackage{multirow}
\usepackage{flushend}
\usepackage{MnSymbol}

\usepackage{caption}

\def\tablename{Table}
\DeclareCaptionLabelFormat{mytablestyle}{\tablename~\thetable}
\renewcommand{\thetable}{\arabic{table}}

\graphicspath{ {./images/} }

\makeatletter
\def\convertto#1#2{\strip@pt\dimexpr #2*65536/\number\dimexpr 1#1}
\makeatother


\title{\LARGE \bf
	Systematically Exploring the Landscape of Grasp Affordances via Behavioral Manifolds
}

\author{Michael Zechmair$^{1}$ and Yannick Morel$^{1}$
    \thanks{$^{1}$Michael Zechmair and Yannick Morel are with Faculty of Psychology and Neuroscience, Maastricht University,
	   Maastricht, The Netherlands,
	   {\tt\small \{m.zechmair,y.morel\}@unimaas.nl}}%
    \thanks{Work partially supported by the European Union's Horizon 2020 Framework Programme for Research and Innovation under the Specific Grant Agreement No. 785907 (Human Brain Project SGA3).}%
}

\begin{document}
	\newcolumntype{a}{>{\columncolor{lightgray}}c}

	\maketitle
	\begin{abstract}
The use of machine learning to investigate grasp affordances has received extensive attention over the past several decades. The existing literature provides a robust basis to build upon, though a number of aspects may be improved. Results commonly work in terms of grasp configuration, with little consideration for the manner in which the grasp may be (re-)produced from a reachability and trajectory planning perspective. In addition, the majority of existing learning approaches focus of producing a single viable grasp, offering little transparency on how the result was reached, or insights on its robustness. We propose a different perspective on grasp affordance learning, explicitly accounting for grasp synthesis; that is, the manner in which manipulator kinematics are used to allow materialization of grasps. The approach allows to explicitly map the grasp policy space in terms of generated grasp types and associated grasp quality. Results of numerical simulations illustrate merit of the method and highlight the manner in which it may promote a greater degree of explainability for otherwise intransparent reinforcement processes.
%
	\end{abstract}

	\begin{figure*}
		\frame{\includegraphics[width=\textwidth]{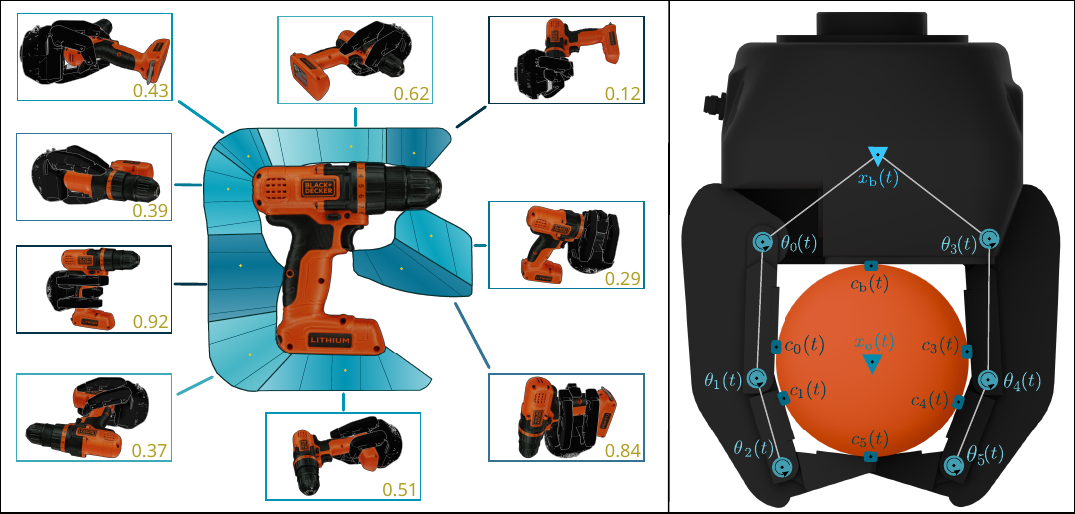}}
		\caption{Considered end-effector ({\small\texttt{Robotiq}} 3-digit gripper), shown example behavioral manifolds for a hand-drill (from the YCB dataset) for the considered gripper and grasp synthesis policy discussed in Section \ref{sec:grasp_synthesis} (left); grasping a cylinder, actuated joints, centers of mass of the cylinder and gripper base, and contact points shown in blue (right).}%
		\label{fig:grippergraspball}%
	\end{figure*}%
	\section{Introduction}\label{sec:introduction}%
The use of machine learning in robotic automation is dramatically impacting artificial agents' capabilities, accelerating learning and skill development capacities, providing perspectives of productive artificial cognitive systems in use on factory floors (\cite{golovianko2023industry}). A meaningful dimension in these developments is the growing use of Reinforcement Learning (RL) to support robotic manipulators in learning to act upon objects of interest (\cite{yang2023recent}). Such advances are of special import in the perspective of developing adaptable, rapidly-reconfigurable production lines. Industrial processes focusing on low-volume production of customized products have received increasing attention (\cite{schuh2017}). Such specialized manufacturing methods require the handling of components of varying geometry and inertia. In practice today, the adjustment of production lines oftentimes remains the responsibility of specialized industrial engineers, providing expert knowledge. Such a process typically proves effort-intensive, motivating the investigation of technologies supporting its automation. A large portion of operations involved in the considered manufacturing processes require the use of a robotic arm equipped with an end-effector, supporting applications that include pick and place, tool manipulation, assembly or disassembly. These tasks commonly require that the end-effector be capable of effectively grasping objects of interest, which has motivated investigations in the area of grasp affordance learning (\cite{kokic2017affordance}). Results pursued are typically characterized in terms of a desirable distribution of contact points or surfaces between grasped object and robotic gripper (or equivalent representation, referred to as \textit{grasp configuration}), affording a number of desirable qualities (e.g. in terms of resilience to perturbations). Work in this area has been structured along two distinct avenues of investigation, analytical and empirical approaches. 

Analytical methods analyze properties of the target object, gripper, and task parameters to determine a grasp using a predefined algorithm. Commonly, such methods pursue emergence of force closure grasps, in which the grasp configuration allows the gripper to apply force on the object in all directions of motion (\cite{bicchi1995closure}). A grasp satisfying this criterion is intended to allow application of counteracting force against external efforts (gripper actuation system's capabilities allowing). In \cite{nguyen1988constructing}, the authors demonstrate an approach allowing to synthesize grasps approximately satisfying closure properties for simple shapes such as polygons. The authors of \cite{objects1993finding} expand on these concepts to find all antipodal point-pairs on arbitrary 3D-objects, which can be used to synthesize grasps for prismatic grippers. This concept was expanded upon in an iterative manner, culminating in an approach capable of finding grasps containing multiple contact points, while still satisfying force closure grasp constraints (\cite{zhu2003synthesis,roa2008independent,roa2009computation}). Recent results expand on these techniques by accounting for additional such constraints. In \cite{berenson2008grasp}, the authors present an approach that classifies resulting grasps based on their efficacy in a cluttered environment, favoring grasps allowing to avoid collisions with nearby objects. In another extension on the approach, the result in \cite{abdeetedal2018grasp} demonstrates an ability to generates grasps capable of applying mechanical stress to objects, with applications to agricultural harvests (e.g. plucking fruits from trees).

Empirical methods typically emulate previously established grasp strategies. This is commonly pursued by recording human grasps, and applying RL techniques to emulate them with the considered robotic system. The range of possible grasps are defined in an established taxonomy (\cite{feix2009comprehensive}). In \cite{zhang2018deep}, the authors use a virtual reality setup to train a robot to imitate human behavior during task execution. Supervised learning has also been used to support grasp synthesis, either by observing  and reproducing successful grasps (\cite{abbeel2004apprenticeship}), or via an annotated mix of failed and successful grasps (\cite{xie2019learning}). Recent approaches use meta-learning to pursue one-shot imitation capabilities (\cite{mandi2022towards, finn2017one}) from visual input. These models are trained using general grasp training sets (such as those in \cite{mahler2018dex,fang2020graspnet}), then fine-tuned via demonstration. Other empirical approaches have developed grasp synthesis strategies by applying reinforcement learning techniques to physical setups featuring robotic arm, end-effector, and a series of objects of interest. The approach amounts to repeatedly attempting to grasp an object, and iteratively improving grasping behavior. Other empirical approaches rely on template matching, whereby a set of template grasps are defined for a considered set of objects. When a novel object is encountered, its shape is analyzed, similarities to known shapes identified, and an appropriate template grasp is selected. This is typically pursued by either mapping template reference frames of grasps to new objects (\cite{morales2006anthropomorphic,do2009grasp}), or by simulating a considered grasp before execution (\cite{morales2006anthropomorphic}). Recently, new developments have focused on artificial expansion of existing image datasets for zero-shot learning. In \cite{vuong2023grasp}, the authors use a large language model, coupled with a latent diffusion model, to generate a large dataset of objects for grasping. The images are automatically annotated to identify viable grasps using a prismatic gripper. While this approach does create a large dataset, the annotations lack real-world information about the included objects; the generated images lack depth, and vital parameters of created objects such as mass, inertia, and material friction coefficients are missing.

Existing approaches suffer from a number of limitations. Quality metrics relied upon to support learning typically reflect a priori expertise, e.g. reproducing human grasps (\cite{zhang2018deep, mandi2022towards, finn2017one}) or satisfying a range of criteria (e.g. force closure, \cite{zhu2003synthesis,roa2008independent,roa2009computation}). The rationale followed is sound, as such criteria have proved to be conducive to the emergence of reliable grasping configurations. However, the approach may suffer from a lack of generalizability in situations where the considered combination of gripper and object does not conform with the a priori knowledge. More importantly, a posteriori assessment of grasp quality is typically limited to whether or not the emerging grasp is viable, with little insights into its practical resilience (\cite{newbury2023deep}). The focus on grasp viability as a binary notion has additional adverse implications. In particular, to promote grasp viability, approaches oftentimes maximize end-effector joint torques or axis efforts (\cite{liu1999qualitative}), which is not necessarily desirable in practice, and may also color the outcome of the learning process. Further, the learning process in most instances is limited to proposing a single grasp configuration, with no information on the range of possible alternative grasps, or insights in terms of the rationale followed to reach the result (\cite{newbury2023deep}), leading to a deficit of transparency and explainability (typical to RL approaches, \cite{lu2023closer}). Finally, results in the literature consider the problem from the perspective of \textit{grasp configuration}. Establishing the range of such possible configurations for simple object geometries (e.g. sphere, cube) and gripper kinematics (prismatic gripper) is straightforward (\cite{bicchi1995closure,nguyen1988constructing,berenson2008grasp,abdeetedal2018grasp}). However, for more complicated object geometries and grippers, analytically establishing viable solutions rapidly becomes intractable.

Other approaches focus on more complex object geometries. In \cite{mordatch2012discovery, mordatch2012contact}, the authors propose a method of contact-invariant optimization that enables the generation of grasp trajectories designed to achieve a specific task. Later on, Grasp'd  adapts this approach to generate grasps for a given gripper-object pair (\cite{turpin2022grasp}). An alternative approach to grasp generation is presented by the authors in \cite{lundell2021multi}. They propose Multi-fingan, an algorithm to generate grasps from RGB-D information. By training a classification network on manually generated grasps, they are able to rapidly infer new grasps for presented objects. However, these approaches aim to find and classify a single grasp for a given gripper-object pair. In case of a changed scenario, the grasp must be regenerated.

To address the aforementioned issues, we propose a generative approach to learning grasp affordances. We rely on the existence of a \textit{grasp synthesis} policy which, given an admissible relative pose between gripper and object, defines what trajectories are applied to the gripper's degrees of freedom (gripper base and actuated axes or joints) in pursuit of grasp emergence. Such a policy $\mathcal G$ provides a map from relative pose space (denoted $\mathcal P$ in the following) to grasp configuration space $\mathcal C$; that is, $\mathcal G: \mathcal P \rightarrow \mathcal C$. Relying on the existence of such a map, one is able to approach the problem of learning how to grasp an object by formulating grasp policies over the relative pose space $\mathcal P$, rather than the configuration space $\mathcal C$, which affords a number of benefits. Under consistent grasp policy, well-defined sets in $\mathcal P$ can be found to lead to consistent grasp types or behaviors. Establishing the landscape of such \textit{behavioral manifolds} (i.e. sets of points $p\in \mathcal P$ mapping to consistent grasp types in $\mathcal C$ under $\mathcal G$, see example in Fig. \ref{fig:grippergraspball}) provides an explicit representation of the range of grasp solutions for the gripper, object, grasp policy tuple. Additionally, we propose a systematic method of exploring grasp configuration space via relative pose space. Using this approach, we can quickly generate a concise mapping that describes regions with similar grasp configurations. This enables us to quantify and compare the performance of reachable grasp configurations.

In the following, we explore the merit of the proposed approach. In Section \ref{sec:grasp_affordance_metric}, we discuss the use of a previously introduced dynamic quality metric (\cite{zechmair2021assessing}) to assess grasp resilience during object manipulation. Then, in Section \ref{sec:grasp_affordance_metric}, we propose a simple example of grasp synthesis policy, and use it to establish the behavioral landscape in relative pose space, for a number of example objects. We further assess impact on performance of informing grasp synthesis using haptic information, and evaluate impact in terms of the size of behavioral manifolds. In Section \ref{sec:grasp_space_exploration}, we showcase our proposed method for systematically exploring the relative pose space to efficiently find and categorize regions with similar grasp configurations. Section \ref{sec:ee_comparison} compares our approach to existing end-to-end learning methods.
Section \ref{sec:conclusion} concludes this paper.

    \begin{figure*}[h]
    	\includegraphics[width=\textwidth]{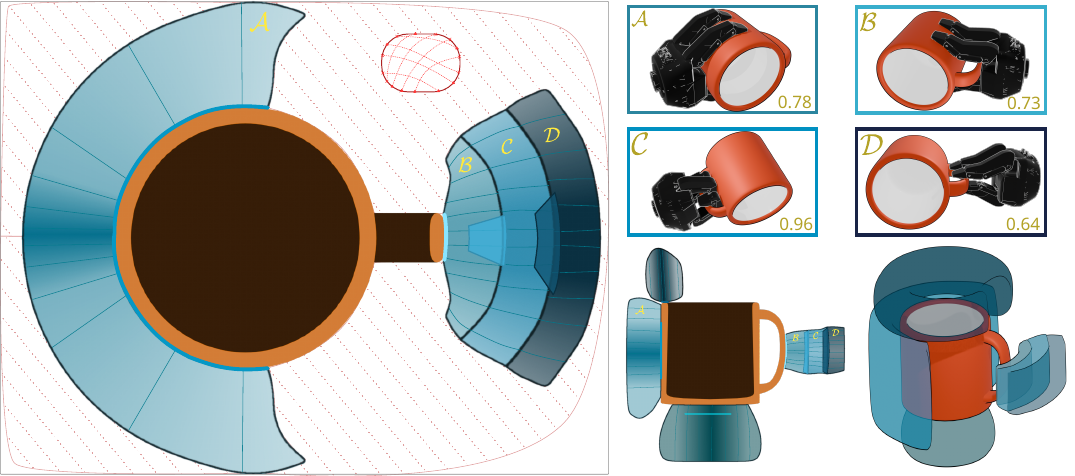}
    	\caption{Horizontal cut of behavioral map for a cup (left), grasped with a {\small\texttt{Robotiq}} 3-fingered gripper, behavioral manifolds $\mathcal{A}$ through $\mathcal{D}$ map to consistent grasp behaviors (right). Contours inside manifolds show $\mu_{\rm s}$ level sets, local minima marked with yellow dots. The red line shows an example optimization trajectory from PPO (jumps in light green). Vertical cut shown bottom right, 3D perspective bottom far right. The behavioral map provides a holistic perspective of possible grasps.}
    	\label{fig:grasp_synthesis_sets}
    \end{figure*}%
	\section{Assessing Grasp Quality} \label{sec:grasp_affordance_metric}%
\begin{table}
	\centering
	\begin{tabular}{|l||c|c|c|c|}
		\hline
		& $\min\tau_{\rm s}$ & $\min\tau_{\rm g}$ & $\min\tau_{\rm f}$ & $\min\tau_{\rm m}$ \\
		\hline\hline
		\textbf{Simple Shapes} & & & & \\
		\hline
		Sphere & 1.48Nm & \textbf{1.48Nm} & 1.48Nm & 1.48Nm     \\
		Dumbbell & \textbf{1.29Nm} & 1.53Nm & 1.29Nm & 1.31Nm    \\
		Pringles Can & \textbf{1.54Nm} & \textbf{1.54Nm} & 1.55Nm & \textbf{1.54Nm} \\
		\hline
		\textbf{Small Objects} & & & & \\
		\hline
		Spoon & 0.83Nm & 0.84Nm & \textbf{0.81Nm} & 0.85Nm \\
		Fork  & 0.83Nm & 0.84Nm & \textbf{0.82Nm} & 0.85Nm \\
		Screwdriver & \textbf{0.93Nm} & 0.99Nm & 1.00Nm & 0.96Nm \\
		\hline
		\textbf{Uneven Mass}   & & & & \\
		\hline
		Power Drill & \textbf{1.87Nm} & 2.13Nm & 2.31Nm & 2.03Nm \\
		Hammer & 2.32Nm & 2.32Nm & \textbf{2.30Nm} & 2.33Nm \\
		\hline
		\textbf{Delicate Objects} & & & & \\
		\hline
		Banana & \textbf{0.52Nm} & \textbf{0.52Nm} & 0.53Nm & \textbf{0.52Nm} \\
		Tomato & \textbf{0.51Nm} & 0.53Nm & 0.53Nm & \textbf{0.51Nm} \\
		Foam Ball & \textbf{0.20Nm} & \textbf{0.20Nm} & \textbf{0.20Nm} & \textbf{0.20Nm} \\
		\hline
		\textbf{Complex Geometry} & & & & \\
		\hline
		Stanford Bunny & \textbf{0.88Nm} & 1.18Nm & 1.24Nm & 1.15Nm \\
		Spiked Ball & \textbf{1.46Nm} & \textbf{1.46Nm} & \textbf{1.46Nm} & \textbf{1.46Nm} \\
		\hline
		\textbf{Slippery Objects} & & & & \\
		\hline
		Wet Sphere & 1.94Nm & 1.94Nm & \textbf{1.93Nm} & 1.94Nm \\
		\hline
	\end{tabular}
	\caption{Minimum joint efforts required to maintain a generated grasp along a trajectory. We compare the grasps generated using the metric defined in Section II ($\tau_{\rm s}$), grasp matrix eigenvalues ($\tau_{\rm g}$), force closure grasps ($\tau_{\rm f}$), and $\tau_{\rm m}$.}
	\label{table:metric_force_comp}
\end{table}%
In this Section, we briefly summarize the dynamic grasp metric first introduced in \cite{zechmair2021penalty}, qualitatively discuss its merit in comparison to static metrics, and explore its application to objects in the YCB data set (\cite{calli2017yale}).

\subsection{Dynamic Grasp Metric}
Grasp metrics characterize the quality of considered grasps, assessing their ability to endure adverse factors without releasing the object. Robotic grasping being commonly involved in manipulation tasks, accounting for eventualities occurring during task performance (e.g. movements undergone by the object in pick and place) is desirable. However, the majority of metrics in the literature characterize grasp quality in a static setting (see \cite{roa2015grasp,abdeetedal2018grasp}), and do not account for factors emerging dynamically. To better account for such aspects, we investigate the application of the dynamic grasp quality metric in \cite{zechmair2021assessing} to support learning of grasp affordances. The method used to derive the metric relies on notions of local sensitivity analysis (\cite{morio2011global}), characterizing interactions between gripper and object accelerations. The formulation requires a closed-form expression of contact dynamics, which we obtain using the physics-based digital contact model in \cite{zechmair2021penalty}. Let $\eta_{\rm o}(t)$, $\eta_{\rm r}(t) \in \mathds{R}^6$, $t\geqslant0$, describe the object's pose and that of a reference point on the gripper base. Movements of the grasped object can be described by
\begin{align}
\begin{split}
	\ddot{\eta}_{\rm o}(t) = %
\begin{bmatrix}
	 f_{\text{g}}(t)/m_{\rm o} + g\\
	I^{-1}_\text{o} \tau_\text{g}(t) - 2 I^{-1}_\text{o}\Omega_{\text{o}}(t) I_\text{o} \omega_{\text{o}}(t)
\end{bmatrix}, \label{eqn:ang_eom}
\end{split}
\end{align}%
where $m_{\rm o} \in \mathds{R}$ is the object's mass in kg, $I_{\rm o} \in \mathds{R}^{3 \times 3}$ its rotational inertia matrix in kg.m, $g \in \mathds{R}^3$ is gravity's acceleration in m/${\rm s}^2$,
$f_{\text{g}}(t)$, $\tau_\text{g}(t) \in \mathds{R}^3$ describe impact of gripper-object interaction forces on the object's dynamics (in N and N.m, respectively). 
Consider the following partial,
\begin{align}
\begin{split}
\frac{\partial \ddot{\eta}_{\rm o}}{\partial \ddot{\eta}_{\rm r}} = \sum_{i=0}^{n} \frac{\partial \ddot{\eta}_\text{o}}{\partial f_\text{g}} \frac{\partial f_\text{g}}{\partial f_{\text{g}i}} \frac{\partial f_{\text{g}i}}{\partial \partial \ddot{\eta}_i}%
\frac{\partial \ddot{\eta}_i}{\partial \ddot{\eta}_{\rm r}},
\end{split}
\end{align}
where $f_{\text{g}i} \in \mathds{R}^3$ is the contact force for the $i^{\rm th}$ contact point, $i=1$, $\ldots$, $n$, and $\ddot{\eta}_i(t) \in \mathds{R}^6$ is the pose acceleration of the $i^{\rm th}$ gripper link (see \cite{zechmair2021penalty, zechmair2021assessing} for details). Evaluating the above partial at time instant $t_\text{x}$, we have
\begin{gather}
G(t_\text{x}) \triangleq
\left. \frac{\partial \ddot{\eta}_\text{o}(t)}{\partial \ddot{\eta}_{\rm r}(t)} \right|_{t=t_\text{x}}, \quad -\epsilon \leqslant t - t_\text{x} \leqslant \epsilon, \nonumber
\end{gather}
where $G(t_\text{x}) \in \mathds{R}^{6 \times 6}$ characterizes the impact of the gripper's movements on those of the object. Minimizing relative movements between gripper and object (i.e. favoring stable relative pose) amounts to minimizing $S(t) = G(t) - I_6$, where $I_6$ is the 6-dimensional identity. Matrix $S(t)$ characterizes the extent to which the considered grasp mitigates the impact of the gripper's movements onto the relative motion of the grasped object (see discussion of robustness to perturbations in \cite{morio2011global}).
%
%
Resilience of gripper movements along independent axes of motion can be assessed by singular value decomposition of $S(t)$, $\sigma_S(t) \in \mathds R^6$.%
		%
Specifically, we measure grasp quality using 
\begin{equation}
	\mu_{\rm s} = \min\big( 1- \max_{t_0 \leqslant t < t_{\rm t}}(\sigma_{S}(t))\big),
\end{equation}%
where $t_0$, $t_{\rm t} \in \mathds{R}^+$, are the task's start and end times. Greater values of $\mu_{\rm s}$ correspond to better quality grasps, in terms of stability of the grasped object relative to the gripper's base.

\subsection{Dynamic Metric for Grasp Learning}
Static approaches (\cite{roa2015grasp}) assess quality in light of a priori, fixed criteria, but do not account for the resulting behavior of the gripper-object pair during task performance. Approaches relying on such quality metrics for grasp learning may in some instances achieve undesirable results, see for illustration the example shown in Fig. \ref{fig:force_metric_comparison}. Vignettes in the upper-left show learned grasps for a rectangular plate by a three-digit gripper, the plot represents grasp quality (static and dynamic) as a function of maximum joint efforts. Grasp learning was conducted using Proximal Policy Optimization (PPO, \cite{wang2020truly}). The learning process is repeated, enforcing different maximum joint efforts. The grasp learned using the static metric remains consistent for all allowed joint efforts (see upper row, grasps a, b). Instead, the grasp learned using metric $\mu_{\rm s}$ changes as a function of allowed efforts. For larger efforts (to the right of the effort threshold shown in orange), the grasp learned is identical to that obtained used the static metric (grasp d). Below this threshold, i.e. when lower efforts are available, the process learns to better balance the plate, gripping it by the middle of its long side, using the double-digit side of the gripper below the plate (grasp c). 
In the following, we explore how the proposed dynamic metric allows to learn effective grasps, for objects from the YCB dataset, requiring smaller joint efforts.
		\subsection{Application to YCB Dataset}
		Note that, as illustrated by the example in \ref{fig:force_metric_comparison}, increasing joint efforts commonly lead to improving value of grasp quality metrics. Accordingly, when such metrics are used to support grasp learning, they do not necessarily emphasize exploration of alternate, more advantageous contact-point locations (or grasp configurations), but may emphasize the use of increasing joint efforts. This may lead to the unintended consequence of requiring unnecessary efforts, and failing to recognize desirable grasp configurations (providing good grasp resilience, without requiring excess efforts). We investigate this notion by using a range of different grasp metrics to learn grasp affordances, for the same gripper as that shown in Fig. \ref{fig:grippergraspball}, applied to a range of objects in the YCB dataset. Then, we investigate resilience of learned grasps by reducing maximal allowed joint torques until we reach grasp failure (i.e. release of grasped object when traveling along a reference trajectory). Results are shown in Table 1 for three static grasp metrics $\mu_{\rm g}$, $\mu_{\rm f}$, and $\mu_{\rm m}$, along with the dynamic metric introduced in the previous Subsection, $\mu_{\rm s}$. The metric $\mu_{\rm g}$ maximizes grasp matrix eigenvalues as described in \cite{li1988task}, $\mu_{\rm f}$ analyzes the object's shape and finds the grasp that maximizes the encompassing grasp volume spanning all contact points (\cite{roa2015grasp}), $\mu_{\rm m}$ uses the method described in \cite{kim2001optimal} to find the grasp maximizing external perturbation resistance. For simpler geometries (sphere, cylindrical can) for which there is no special grasp affordance to learn, results are even across metrics. However, in situations where special grasp configurations may be found to promote stability, the proposed dynamic metric results in grasps resilient to lower joint torques. This is for instance the case for the hand drill in Fig. \ref{fig:grippergraspball}, where the grasp configuration around the handle, leaning against the upper part of the body, is only learned using the proposed dynamic metric, and proves more resilient than solutions reached using alternate metrics.
%

 
	\section{Behavioral Map}\label{sec:grasp_synthesis}%
	The grasp configuration space $\mathcal C$ describes possible configurations of the gripper and object when contact occurs, and ideally, a viable grasp is formed. Definition of this space is non-unique; in some instances authors have used 
SE(3) (\cite{gualtieri2016high,ten2017grasp,liang2019pointnetgpd}), though it proves only sufficient to describe the gripper's pose relative to the object. In situations in which gripper axes configuration may be unequivocally inferred from relative pose $\eta_{\rm r}$ and object geometry (e.g. as for a prismatic gripper making contact with a sphere), it may prove sufficient to characterize grasp configurations. In a general setting however, expanding the space to include gripper axes or joints coordinates $\theta_{\rm g}$ may prove necessary, leading to the following possible definition of the configuration space,   
%
\begin{equation}
	\mathcal{C} = \left\{\; \eta_\text{r}, \theta_\text{g} \; | \; n_\text{c} > 0 \right\},
\end{equation}%
where $n_\text{c}$ is the number of contact points between gripper and object. Working in terms of such a set is however nontrivial. In particular, dimension of $\mathcal C$ may be large, depending on number of actuated degrees of freedom of the gripper, the contact constraint is not necessarily easy to work with (or even describable in a closed form for non-trivial shapes), parsing and visualizing viable regions of $\mathcal C$ (i.e. points in $\mathcal C$ that correspond to viable grasps), representing regions of $\mathcal C$ corresponding to better or worse grasp quality metrics, can prove less than intuitive (see one of the few attempts in \cite{aleotti20123d}).
\subsection{Behavioral Manifolds in Pose Space}
We instead propose to approach the grasp affordance learning problem from the perspective of the relative pose space
%
\begin{equation}
	\mathcal{P} = \left\{\; \eta_\text{r} \; | \; d_\text{go} \leqslant d_\text{p} \right\},
\end{equation}%
where $d_\text{go} \in \mathds{R}^+$ is the distance between gripper base and object surface, and $d_\text{p} \in \mathds{R}^+$ is a threshold distance defining the exploration radius around the object. We in addition consider a grasp synthesis policy $\mathcal G$, characterizing trajectories prescribed to the gripper base and actuated DoFs, intended to promote emergence of viable grasps, which operates as a mapping from relative pose to configuration space,
%
\begin{equation}
	\mathcal{G}: \mathcal{P} \rightarrow \mathcal{C}.
\end{equation}%
With minimal requirements on $\mathcal G$, $\mathcal P$ is mapped to a subset of $\mathcal C$ under $\mathcal G$. Different such grasp synthesis policies are discussed in latter Subsections.

In practice, one is able to explore the relative pose space $\mathcal P$ and investigate outcome for different points and regions of $\mathcal P$ under mapping $\mathcal G$. We disregard points in $C$ corresponding to non-viable grasps (i.e. contact is made between part(s) of the gripper and part(s) of the object, but no effective grasp is formed), and instead focus on regions of $\mathcal P$ leading to the formation of viable grasps under $\mathcal G$. Under straightforward grasp synthesis policies, we observe the emergence of well-defined, compact (closed, bounded) sets in $\mathcal P$ leading to consistent grasp types. Outcome of such an exploration is represented in Fig. \ref{fig:grasp_synthesis_sets}.
Specifically, the diagram on the left of Fig. \ref{fig:grasp_synthesis_sets} shows a sample surface of $\mathcal P$ corresponding to the horizontal plane (with coordinates representing the two-dimensional position of a point of reference on the base of the gripper, gripper attitude constrained such that the gripper base is parallel to the nearest tangent of the mug body, dual-finger side of the gripper on the upper side). The object considered is a mug, the gripper is that in Fig. \ref{fig:grippergraspball}, the grasp synthesis policy that in Section III.C. Lightly dashed areas correspond to regions of space not leading to viable grasps, sets labeled $\mathcal A$ through $\mathcal D$ map to different grasp types; force closure around the cup body from $\mathcal A$, pinching grasp around the cup body from $\mathcal B$, force closure around the cup handle from $\mathcal C$, pinching grasp on the handle from $\mathcal D$ (grasp types shown in Fig. \ref{fig:grasp_synthesis_sets}, top right). Emerging grasp behaviors, from each labeled set, is consistent under $\mathcal G$. We accordingly refer to them as \textit{behavioral manifolds}. Grasp quality, as assessed by the metric in Section II, marginally varies within each set, color gradients in Fig. \ref{fig:grasp_synthesis_sets} illustrate metric gradients, level sets are represented in solid lines. Local extrema are marked with yellow dots. A projection of sets in $\mathcal C$ that behavioral manifolds map to under $\mathcal G$ are also represented in Fig. \ref{fig:grasp_synthesis_sets} (showings the position of a reference point of the gripper surface in the horizontal surface represented). Set $\mathcal A$ maps to the 1-dimensional set along the surface of the cup represented with a thick solid blue line, $\mathcal B$ maps to the light blue line on the surface of the handle, $\mathcal C$ and $\mathcal D$ to the small 2 dimensional sets located to their left, represented with thinner outlines.

The behavioral map in Fig. \ref{fig:grasp_synthesis_sets} provides a number of interesting features. In particular, it allows to visualize the range of possible grasp outcomes (both grasp type and quality, for a given $\mathcal G$ and metric) in a concise, self-explanatory representation. Further, consideration of behavioral manifolds provides information helpful to planning. Specifically, achieving a desired grasp type only requires reaching the relevant manifold and following the corresponding grasp synthesis. In addition, representing optimization trajectories onto the behavioral map allows to visualize and apprehend the rationale followed by the process. In Fig. \ref{fig:grippergraspball}, the trajectory generated by PPO is represented it red. In initializes in the top left corner, explores a non-viable area before entering $\mathcal A$. It then descends different gradients, localizes a number of local minima, follows the outline of different manifolds, jumps to different regions to escape local minima before eventually converging to a (local) minimum. The map provides explainability and transparency to the RL process.

%
\begin{figure}
	\centering
	\def\svgwidth{\columnwidth}
	\includegraphics[width=\columnwidth]{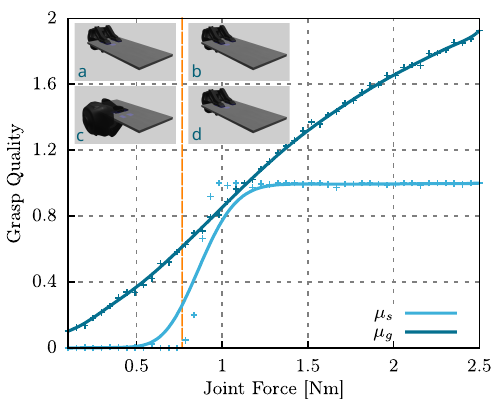}%
	\caption{Static ($\mu_{\rm g}$) and dynamic ($\mu_{\rm s}$) grasp metrics as a function of maximum joint torques, grasping a rectangular plate. Grasps learned using the static metric in the top row (grasps a, b), dynamic metric in the bottom row (c, d).}
	\label{fig:force_metric_comparison}
\end{figure}%
		\subsection{Baseline Grasp Synthesis Policy}%
Valid grasp configurations require emergence of contact between designated parts of the gripper and object surface. In the following we consider a simple grasp policy for the three digit gripper shown in Fig. \ref{fig:grippergraspball}. From an initial relative pose at a short distance from the object, gripper base parallel to the nearest surface of the object, digits opened to their widest, such contacts are pursued by approaching the gripper base and digits towards the object's surface. In this perspective, we define a baseline grasp synthesis policy $\mathcal{G}_\text{c}$. The policy relies on knowledge of the pose $\eta_\text{r}(t_\text{p})$ of the gripper's CoM relative to the object's CoM at initial time instant $t_\text{p}$. The policy then assesses (from the object's geometry) the point on the object's surface $x_\text{c}$ that is closest to the gripper base's surface. Then, a constant force of magnitude $f_\text{c}$ and direction defined as that from gripper CoM to $x_\text{c}$ is applied to the gripper. As contact is made between gripper and object, the force is removed, and a constant closing torque $\tau_\text{c}$ is applied to gripper joints, until the time instant $t_\text{c}$ at which gripper movements have halted (under resistance from the object). The resulting gripper base pose, relative to that of the object, appended with the gripper's joint angles, $\left( \eta_\text{r}(t_\text{c}), \theta_\text{g}(t_\text{c}) \right)$ describe the grasp configuration. The proposed policy thus describes the following map,
\begin{eqnarray}
	\mathcal G_{\rm{c}}: \quad \quad \mathcal P &\rightarrow& \mathcal C, \nonumber \\ 
                        \eta_{\rm r}(t_{\rm p}) &\rightarrow& \eta_{\rm r}(t_{\rm c}),\; \theta_{\rm g}(t_{\rm c}).
\end{eqnarray}
In the following, we adjust the previously discussed policy to exploiting proximity information made available from a haptic sensor such as that described in \cite{zechmair2023active}.%
		\subsection{Haptic Grasp Synthesis Policy}%
\begin{figure}
	\centering
	\includegraphics[width=\columnwidth]{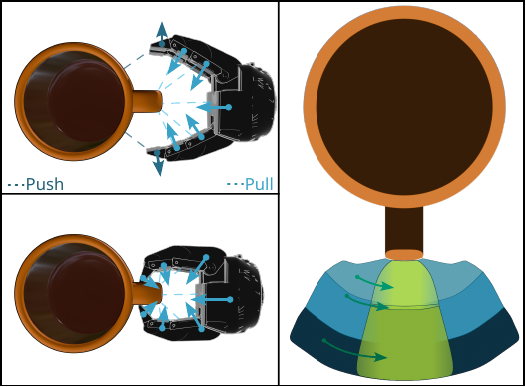}
	\caption{Proximity perception is used to conform shape of the gripper to that of the object, links in closer proximity move away, those further away move closer (left). Exploiting proximity information expands behavioral manifolds (right).}%
	\label{fig:shape_adaptation_comparison}
\end{figure}%
\RestyleAlgo{boxed}
\LinesNumbered
\SetAlFnt{\footnotesize}
\IncMargin{0.4em}
\begin{algorithm}[t!]
	\DontPrintSemicolon	
	\SetKwProg{Fn}{Function}{}{}
	\Fn{GraspSynthesis($\eta_{\rm r}$, $\theta_{\rm g}$, $f_{\rm s}(\cdot)$)}
	{
		\tcc{Adjust digits so that distance to object is within tolerance}
		\While{$\max \hat{d}_{{\rm s}i} > \hat{d}_{\rm a}$}{%
		\tcc{Jacobian (\ref{equ:alignment_jacobian})}
		$\hat{J}_{\rm s}$ = est\_finger\_jacobian($\hat{d}_{{\rm s}}$, $\eta_{\rm r}$, $\theta_{\rm g}$)\\
		$\theta_{\rm g} \leftarrow \hat{J}_{\rm s}^+ \theta_{\rm d} + \theta_{\rm g}$\\
		\tcc{Update digit kinematic chains}
		\ForEach{$i \in n$}{%
		$\eta_i = k_i(\eta_{\rm r}, \theta_{\rm g})$
		}}%
		\BlankLine
		\BlankLine
		\BlankLine
		\BlankLine
		\tcc{Reduce distance to object in a homogeneous manner for  grasp emergence}
		\While{$\max \hat{d}_{{\rm s}i} > \hat{d}_{\rm c}$}{%
			\tcc{Jacobian from \ref{sec:sa_2_grasp_closure})}
			$J_{\rm s}$ = est\_gripper\_jacobian($\hat{d}_{{\rm s}}$, $\eta_{\rm r}$, $\theta_{\rm g}$)\\
			$\begin{bmatrix}
				\eta_{\rm r}\\
				\theta_{\rm g} 
			\end{bmatrix} \leftarrow \hat{J}_{\rm s}^+ %
			\begin{bmatrix}
				\eta_{\rm d}\\
				\theta_{\rm d}
			\end{bmatrix} + %
			\begin{bmatrix}
				\eta_{\rm r}\\
				\theta_{\rm g}
			\end{bmatrix}$\\
			\tcc{Update gripper kinematic chain}
			\ForEach{$i \in b, [0, n)$}{%
				$\eta_i = k_i(\eta_{\rm r}, \theta_{\rm g})$
		}}%
	}
	
	\BlankLine
	\BlankLine
	\BlankLine
	\tcc{Return the grasp configuration}
	\KwResult{
		$\begin{bmatrix}
			\eta_{\rm r}\\
			\theta_{\rm g}
		\end{bmatrix}$}
	
	\caption{Shape-informed grasp synthesis policy, implementing the steps outlined in \ref{sec:sa_1_digit_alignment} and \ref{sec:sa_2_grasp_closure}.}
	\label{alg:shape_adaptation}
\end{algorithm}%

The previous grasp policy relied on knowledge of the relative pose of the object's CoM. Hereafter, we instead adjust movements of the gripper base and digits as they approach the object, in a manner that they conform with its perceived local shape.
%
%
The approach is divided in two steps. Step \texttt{1} aims to adjust gripper digits in such a manner that the distance between the surface of individual gripper links and the object's surface nearest to the considered link are homogeneous within a given tolerance $\Delta d_{\rm s} \in \mathds{R}^+$. Step \texttt{2} then adjusts base pose and joint angles to reduce measured distances until a closure grasp is achieved.%
			\subsubsection{Digit Alignment}\label{sec:sa_1_digit_alignment}%
The initial condition assumes all gripper digits are in an open configuration. We evaluate the distance $d_{\rm b}(t)$ between individual links and nearest object surfaces,
\begin{equation}
	d_{{\rm s}i}(t) = f_{\rm s}(\eta_{i}(t), \eta_{\rm o}(t)), \label{equ:surface_dist}
\end{equation}%
where $\eta_{i}(t) \in \mathds{R}^3$ is the pose of link $i$'s CoM, and $f_{\rm s}(\cdot)$ represents a nonlinear function computing the distance. In numerical simulation, $f_{\rm s}(\cdot)$ is provided by typical geometrical libraries, using the respective poses and 3D meshes (\cite{fabri2009cgal}). Specifically, we used {\small \texttt{Bullet}}'s integrated functions. We then define the offset distance
\begin{align}
	\hat{d}_{{\rm s}i}(t) = d_{{\rm s}i}(t) - d_{\rm b}(t), \label{equ:surface_dist_offset}
\end{align}%
where $d_{\rm b}(t) \in \mathds{R^+}$ is the minimum distance between base and considered object. We compute the link poses, 
\begin{align}
		\eta_{i}(t) = k_{i}(\eta_{\rm r}(t), \theta(t)), \label{equ:link_kinematics}
\end{align}%
where $k_{i}(\cdot)$ is the links' kinematics. Substituting (\ref{equ:link_kinematics}) into (\ref{equ:surface_dist_offset}) provides a direct relationship between joint angles and distances. We then consider the Jacobian
%
\begin{align}
	\hat{J}_{\rm s}(t) = \frac{\partial \hat{d}_{\rm s}(t)}{\partial \theta_{\rm g}(t)}. \label{equ:alignment_jacobian}
\end{align}%
Joint angles are adjusted in the direction of $\hat{J}_{\rm s}(t)$ until $\hat{d}_{{\rm s}n}(t)$ and $d_{\rm b}(t_0)$ are within tolerance $\Delta d_{\rm s}$.
If no joint configuration is found fulfilling the requirement, the initial pose is discarded as non-viable ($\mu_{\rm s} = 0$).%
			\subsubsection{Grasp Closure}\label{sec:sa_2_grasp_closure}%
Upon completion of step \texttt{1}, shape adaptation is used to pursue emergence of an enclosure grasp. We compute the Jacobian $J_{\rm s}(t)$ 
as the partial derivative of the distance vector $\begin{bmatrix} d_{\rm b}(t), d_{0}(t), \ldots d_{n}(t) \end{bmatrix}^{\rm T}$ with respect to  base pose and joint angles, which we then adjust in the direction defined by the Jacobian. The process is repeated until a grasp emerges.%
		\subsection{Grasp Policies Comparison}
\begin{figure}
\begingroup%
  \makeatletter%
  \providecommand\color[2][]{%
    \errmessage{(Inkscape) Color is used for the text in Inkscape, but the package 'color.sty' is not loaded}%
    \renewcommand\color[2][]{}%
  }%
  \providecommand\transparent[1]{%
    \errmessage{(Inkscape) Transparency is used (non-zero) for the text in Inkscape, but the package 'transparent.sty' is not loaded}%
    \renewcommand\transparent[1]{}%
  }%
  \providecommand\rotatebox[2]{#2}%
  \newcommand*\fsize{\dimexpr\f@size pt\relax}%
  \newcommand*\lineheight[1]{\fontsize{\fsize}{#1\fsize}\selectfont}%
  \ifx\svgwidth\undefined%
    \setlength{\unitlength}{252.00001526bp}%
    \ifx\svgscale\undefined%
      \relax%
    \else%
      \setlength{\unitlength}{\unitlength * \real{\svgscale}}%
    \fi%
  \else%
    \setlength{\unitlength}{\svgwidth}%
  \fi%
  \global\let\svgwidth\undefined%
  \global\let\svgscale\undefined%
  \makeatother%
  \begin{picture}(1,0.85066006)%
    \lineheight{1}%
    \setlength\tabcolsep{0pt}%
    \put(0,0){\includegraphics[width=\unitlength,page=1]{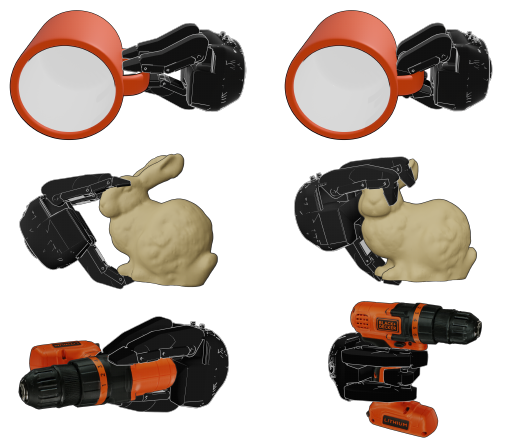}}%
    \put(0.00595238,0.01193691){\color[rgb]{0.4,0.4,0.4}\makebox(0,0)[lt]{\lineheight{1.25}\smash{\begin{tabular}[t]{l}Baseline\end{tabular}}}}%
    \put(0.5058214,0.01218596){\color[rgb]{0.4,0.4,0.4}\makebox(0,0)[lt]{\lineheight{1.25}\smash{\begin{tabular}[t]{l}Shape-aware\end{tabular}}}}%
    \put(0,0){\includegraphics[width=\unitlength,page=2]{interesting_grasps.pdf}}%
  \end{picture}%
\endgroup%

	\caption{Grasps produced by $\mathcal{G}_{\rm c}$ (left) and $\mathcal{G}_{\rm s}$ (right) from the same initial relative pose; $\mathcal{G}_{\rm s}$ results in more resilient grasps.}%
 	\label{fig:grasp_comp}
\end{figure}

Following the baseline grasps synthesis policy, the gripper's digits simply close onto the object after the base makes contact. In situations in which the object's geometry is simple and favorable, the policy proves successful, as shown in Fig. \ref{fig:force_metric_comparison}. 
However, in cases where the object offers a more complex surface geometry, the approach provides mixed results. For instance, small surface features offering opportunities for improved grasps can be overlooked. Fig. \ref{fig:shape_adaptation_comparison} provides an illustrative example, focusing on the opportunity for a stable force closure grasp around a cup's handle (behavior $\mathcal C$ in \ref{fig:grasp_synthesis_sets}). Behavioral manifolds for the baseline grasp synthesis policy are shown in a palette from grey to yellow. The policy allows emergence of both pinching behaviors, although the manifold surfaces are meaningfully reduced in comparison to those achieved using the shape-informed policy (blue palette). More significant, the manifold for handle force closure almost vanishes when using the baseline policy. In other words, in the absence of proximity information, observing emergence of this force closure grasp is unlikely. Similar results were found across different object geometries, as shown in Fig. \ref{fig:grasp_comp}. Specifically, identical relative poses led to substantially different outcomes for either policy, with the shape-informed policy providing higher quality grasps; handle force closure on cup, secure ear grasp on Stanford bunny, handle force closure on the hand drill.
In addition, the emergence of larger behavioral manifolds when using the shape-informed policy was verified over all objects in the YCB dataset. Some results are shown in Table \ref{tab:shape_adaptation_comparison}. In all instances, manifolds obtained from $\mathcal{G}_{\rm s}$ contain those obtained from $\mathcal{G}_{\rm c}$. The average volume increase is of the order of 7\%, objects featuring non-trivial grasp affordances, such as handles, see the greatest benefit.
\begin{table}
	\centering
	\begin{tabular}{|l||c|c|c|c|}
		\hline
		& $v_{\rm c}$ & $v_{\rm s}$ & $\mu_{\rm c}$ & $\mu_{\rm s}$\\
		\hline\hline
		\textbf{Uneven Surface}   & & & & \\
		\hline
		Power Drill    & \textbf{136.6} & 158.3 & \textbf{0.53} &         0.87  \\
		Hammer         & \textbf{138.2} & 163.2 & \textbf{0.74} &         0.76  \\
		Stanford Bunny & \textbf{ 93.5} & 115.5 & \textbf{0.64} &         0.69  \\
		Spiked Ball    & \textbf{113.9} & 125.4 & \textbf{0.43} & \textbf{0.43} \\
		\hline
		\textbf{Simple Shapes} & & & & \\
		\hline
		Sphere       & \textbf{125.7} & \textbf{125.7} & \textbf{0.88} & \textbf{0.88} \\
		Cube         & \textbf{119.2} &         120.3  & \textbf{0.74} & \textbf{0.74} \\
		Dumbbell     & \textbf{138.2} &         139.2  & \textbf{0.94} & \textbf{0.94} \\
		Pringles Can & \textbf{136.2} &         136.5  & \textbf{0.91} & \textbf{0.91} \\
		\hline
		\textbf{Small Objects} & & & & \\
		\hline
		Spoon       & \textbf{93.3} & \textbf{93.3} & \textbf{0.93} & \textbf{0.93} \\
		Fork        & \textbf{93.1} &         93.2  &         0.93  &         0.93  \\
		Screwdriver & \textbf{92.9} &         93.0  &         0.98  &         0.98  \\
		\hline
	\end{tabular}
	\caption{Comparison of manifold volumes $v_{\rm c}$ and $v_{\rm s}$ in $[{\rm cm}^3]$, obtained from $\mathcal{G}_{\rm c}$ and $\mathcal{G}_{\rm s}$, respectively. Metrics $\mu_{\rm c}$ and $\mu_{\rm s}$ are provided for the best obtained grasps.}%
	\vspace{-1em}
	\label{tab:shape_adaptation_comparison}
\end{table}%

	\section{Grasp Space Exploration} \label{sec:grasp_space_exploration}%
	\begin{figure*}
	\includegraphics[width=\textwidth]{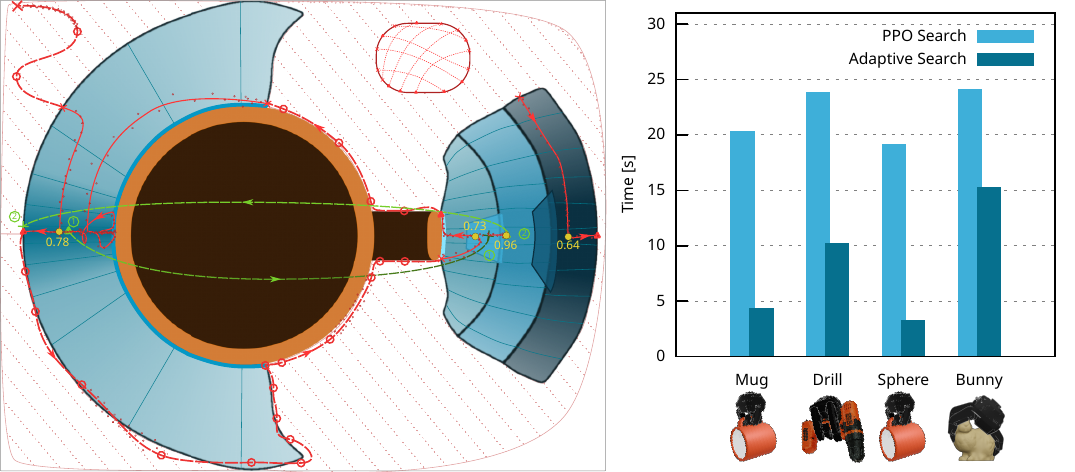}
	\caption{2D-cutout of grasp synthesis regions for a cup, generated with a \texttt{Robotiq} 3-fingered gripper. The red line shows the path Google's PPO-based space exploration algorithm takes, with green arrows denoting jumps of search paths and yellow dots indicating the individual maximas found in each region. The right graph shows a comparison of execution time between PPO and our adaptive search.}%
	\label{fig:ppo_exploration}%
\end{figure*}%
In section \ref{sec:grasp_synthesis}, we describe how to synthesize grasps for a given object. Section \ref{sec:grasp_affordance_metric} introduces a metric to evaluate a grasp's performance.In this section, we will expand on the introduced methods to generate an encompassing set of grasps for an object that covers most approach trajectories, while also ranking said grasps according to their performance. This necessitates a method of exploring grasp space. As previously discussed in section \ref{sec:grasp_synthesis} however, purely searching through all possible grasp configurations requires an unreasonable amount of time. Instead, we explore the lower-dimensional set of relative base poses $\mathcal{P}$ and their mapped grasp configuration set $\mathcal{C}$. This enables us to discard unreachable grasp configurations and focus on a reasonable subset of valid, usable grasps.

\subsection{PPO-based Grasp Space Exploration}
Our initial approach was based on using a Proximal Policy Optimization (PPO)-based reinforcement learning approach, as presented in \cite{schulman2017proximal}. PPO algorithms have previously been used for successful grasp policy learning, as shown by the approaches demonstrated in \cite{fan2018real, zhang2022simulation, pedersen2020grasping, zheng2023object}. We built our setup on top of the AngoraPy framework for goal-oriented research (\cite{weidler2023angorapy}). The framework facilitates a method of integrating various physics simulators and connecting them to reinforcement learning algorithms. For our use-case, we developed a custom physics simulator based on \texttt{Bullet} that allows us to control a gripper, evaluate its interaction with various objects, and compute a grasp affordance metric as presented in Section \ref{sec:grasp_affordance_metric}. AngoraPy then extends an API to connect this simulator with a PPO algorithm to learn grasp policies via reinforcement learning. However, as PPO is a model-free algorithm, we must still provide an our own input and output. Here, our goal is to explore as many grasps as possible for a given object. As stated previously, we are indirectly covering the grasp configuration space $\mathcal{C}$ by exploring the relative pose space $\mathcal{P}$.
Our aim is to determine the most stable grasp while minimizing the employed grasp force.
Taking both goals into account, our input for a single PPO cycle(?) is the initial relative gripper pose $\eta_{\rm rp}$ as well as the maximum allowed gripper force. The output of the simulation is our grasp metric $\mu_{\rm s}$. We configure the simulator's initial conditions by adding both the considered object as well as our gripper at its relative pose. We then run the simulation and synthesize a grasp as described in section \ref{sec:grasp_synthesis}. If the synthesis fails to generate a grasp, we mark the initial pose as invalid and return a score of 0. If the synthesis is successful, we record the resultant grasp configuration $(\eta_{\rm rc}, \theta_{\rm g})$ in our set of valid grasp configurations $\mathcal{C}$. Lastly, we simulate the given trajectory, and return the computed grasp metric score $\sigma_{\rm s}$ (should the grasp fail during execution, the score is 0). Figure \ref{fig:reinforcement_learning_steps} shows an example of the individual steps.

We define the PPO loss function as
\begin{equation}
	L = w_{\rm s} \mu_{\rm s} + w_{\rm f} \tau_{\rm m},
\end{equation}%
where $w_{\rm s} \in \mathds{R}^+$ is the weight assigned to the quality of the evaluated grasp, $\tau_{\rm m} \in \mathds{R}^+$ is the maximum joint torque employed by the evaluated grasp during trajectory execution in Nm, and $w_{\rm f} \in \mathds{R}^+$ is  weight assigned to $\tau_{\rm m}$ in $\frac{1}{\rm Nm}$.

An example 2D-cutout of a PPO-based exploration is shown in Fig. \ref{fig:ppo_exploration}. The search begins at the top left and randomly explores until it finds a region of valid poses. Once achieved, it performs a gradient descent until a local optimum is found, then continues it's search. The approach covers most of the pose space around the mug and finds the pose regions described above. Later, we can group the poses into their respective regions (labeled $\mathcal{A}$ -- $\mathcal{D}$) by looking at their mapped grasp configurations.
With this, we are able to determine a good grasp for a given object within around 20-25 minutes. However, during the development of our approach, we noted a series of properties that could be used to improve search times.
However, PPO is a model-agnostic optimization method. It does not require information about the underlying system. Therefore, while it can be adapted to a multitude of problems without too much effort, it also cannot take advantage of system characteristics while performing a search. To properly use our findings, we had to adapt a separate approach. 

\subsection{Adaptive Grasp Space Exploration}

We want to perform a comprehensive exploration of reachable grasp configurations
PPO-based approach searches for the best grasp using a heuristic approach. We aim to use the object's geometric properties to perform a systematically search. By leveraging our behavioral mapping, we can investigate the simple pose space instead of the more complex and higher-dimensioned configuration space. By exhaustively searching through relative pose space, we can cover a reasonable set of grasp configurations. We propose a succinct search algorithm that covers nearby pose space and focuses on areas of interest (regions where a mapping to a valid grasp configuration exists).
Identify and characterize regions within relative pose space that map to similar grasp configurations. Use spherical coordinates to define the pose of the gripper base's CoM in relation to the object's CoM.
The most important aspect of a mapping is the approach vector. By using spherical coordinates, we can use $\theta, \phi$ to define approach vector and use radius $r$ to adjust distance to nearest object point.
First iterate over both $\theta$ and $\phi$ in steps of $\Delta \theta$.
At each position, adjust $r$ so that the initial distance between the gripper base's surface is at a distance of $r_{\rm i}$  from the closest object point.
After setting the gripper base's relative position, we adjust its attitude so that the gripper base's surface normal is aligned with the normal of the closest surface point on the object.
We also iterate over the same $\theta$, $\phi$, but with a variation in $r$ from $r_{\rm min}$ to $r_{\rm max}$ by $\Delta r$.
At each position, we check whether the pose can be mapped to a valid grasp configuration.
Grasp configurations are identified, and poses that map to similar grasps are added to individual lists.
Using this initial pass-through, we attain a rough outline of possible grasp regions.
Next, we perform a search to determine the border of detected regions.
We check our results for existing poses indicating a nearbyg border of a region (Coordinates where one pose can be mapped to a valid grasp, and a neighboring coordinate without this property).
For each of these coordinates, we performa a search over closer neighboring points with a smaller step size of $\Delta \theta / 2$ and $\Delta r / 2$.
Each new point that is identified to be within the region is added to the list.
We continue iterating over region coordinates until the slopes of line segments between neighboring poses are smaller than $s_{\rm max}$. This enables us to determine regions with arbitrary accuracy.
Within each region, we perform a gradient descent search to find the pose with best grasp quality.

The following lists our observations, then describes how our approach is configured. \textbf{1.} Multiple poses in $\mathcal{P}$ map to the same or very similar grasps in $\mathcal{C}$. If an exploration algorithm is aware of this, it can infer the grasp metric score based on previous evaluation of similar grasps. This would allow us to skip a simulation step, saving time. \textbf{2.} Initial poses along a local normal of an object surface map to the same grasp configuration. Usually, grasps don't change when approaching from the same direction. An example of this is shown in Fig. \ref{fig:ppo_exploration}, in set $\mathcal{A}$. Here, all poses along the object surface normals are mapped to the same grasp. By identifying regions where this is the case, we can skip both grasp synthesis and evaluation along these normals, as all poses will result in the same grasp. \textbf{3.} Exploring objects with shape symmetries result in grasp regions with similar symmetries. For example, the cup shown in Fig. \ref{fig:ppo_exploration} is mirror symmetric along the horizontal axis, and the pose region mirror this symmetry.
Taking these properties into account, we created our own search method, replacing the PPO-based approach; not only does our method speed up the required search time, it also returns more information. Instead of only searching for the most performant grasp, we generate a map of all possible approach trajectories and their corresponding grasp configuration, along with a ranking of each grasp.

\begin{figure}
	\centering
	\def\svgwidth{\columnwidth}
\begingroup%
  \makeatletter%
  \providecommand\color[2][]{%
    \errmessage{(Inkscape) Color is used for the text in Inkscape, but the package 'color.sty' is not loaded}%
    \renewcommand\color[2][]{}%
  }%
  \providecommand\transparent[1]{%
    \errmessage{(Inkscape) Transparency is used (non-zero) for the text in Inkscape, but the package 'transparent.sty' is not loaded}%
    \renewcommand\transparent[1]{}%
  }%
  \providecommand\rotatebox[2]{#2}%
  \newcommand*\fsize{\dimexpr\f@size pt\relax}%
  \newcommand*\lineheight[1]{\fontsize{\fsize}{#1\fsize}\selectfont}%
  \ifx\svgwidth\undefined%
    \setlength{\unitlength}{252bp}%
    \ifx\svgscale\undefined%
      \relax%
    \else%
      \setlength{\unitlength}{\unitlength * \real{\svgscale}}%
    \fi%
  \else%
    \setlength{\unitlength}{\svgwidth}%
  \fi%
  \global\let\svgwidth\undefined%
  \global\let\svgscale\undefined%
  \makeatother%
  \begin{picture}(1,1.11739688)%
    \lineheight{1}%
    \setlength\tabcolsep{0pt}%
    \put(0,0){\includegraphics[width=\unitlength,page=1]{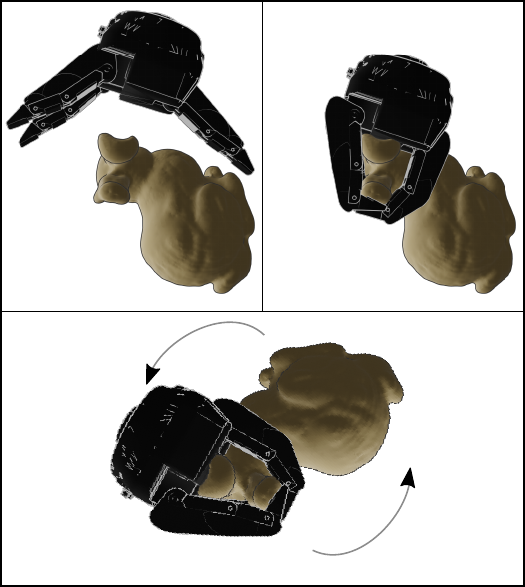}}%
    \put(0.01382447,0.53633558){\makebox(0,0)[lt]{\lineheight{1.25}\smash{\begin{tabular}[t]{l}1. Initialization\end{tabular}}}}%
    \put(0.01190431,0.011906){\makebox(0,0)[lt]{\lineheight{1.25}\smash{\begin{tabular}[t]{l}3. Trajectory Execution and Grasp Analysis\end{tabular}}}}%
    \put(0.51202041,0.53619921){\makebox(0,0)[lt]{\lineheight{1.25}\smash{\begin{tabular}[t]{l}2. Grasp Synthesis\end{tabular}}}}%
  \end{picture}%
\endgroup%

	\caption{Evaluation of a relative pose. Top left image displays the initial gripper configuration, with the gripper base placed at an initial pose relative to the object, top right image shows the synthesized grasp, and bottom image shows the execution of an example trajectory.}
	\label{fig:reinforcement_learning_steps}
\end{figure}
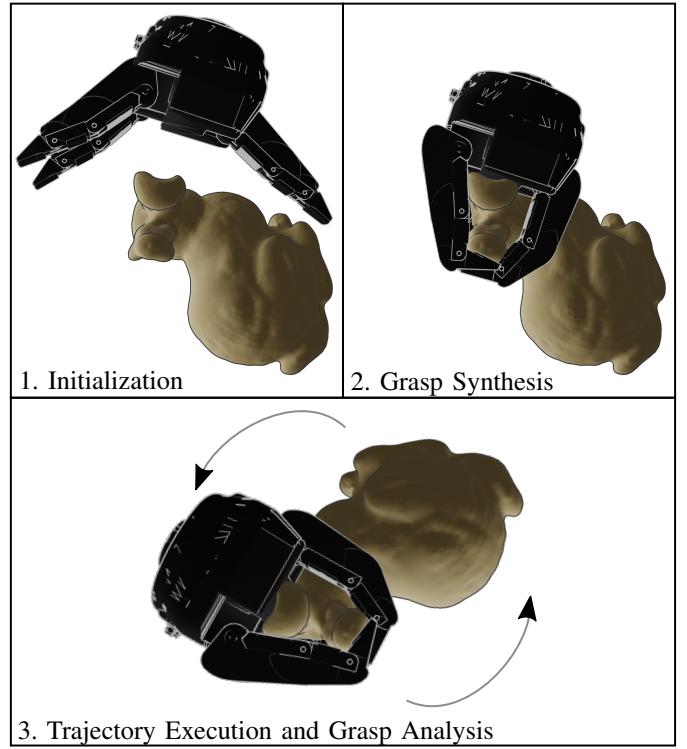

Exploring such a large space in a reasonable amount of time is difficult. Therefore, we developed a method to simplify configuration space exploration; we limit the search to all possible relative base pose, and generate one reasonable joint configuration per pose. Similar approaches have been used in 
. The resulting exploration space is reduced to a 6-dimensional manifold (?) consisting of all possible relative base poses, removing the need to explore multiple joint configurations for each individual pose. The PPO performs a search through this configuration space to find the base pose resulting in the best grasp metric.
In the following, we describe both our approach to generating a reasonable grasp joint configuration for a given relative gripper base pose as well as how we compute the grasp affordance metric for this grasp. See Fig. \ref{fig:reinforcement_learning_steps} for an example of the individual phases of a single grasp evaluation cycle.
\subsection{Performance Comparison}
Comparing both approaches, we note a significant reduction in search times. Examples are shown in the graph of Fig. \ref{fig:ppo_exploration}. For both simple as well as complex shapes, we note a significant reduction in exploration time. Additionally, the adaptive approach yields a complete mapping of nearby pose to grasp configuration space, with each grasp tagged with a corresponding quality metric. PPO-based exploration can yield similar results, but this is not guaranteed; it's main objective is finding the most performant grasp while still sufficiently exploring all possible options.

We evaluated our approach on the YCB dataset and compared it to using a conventional PPO-based approach.
We evaluated our approach while testing various number of attitudes per position (1,2,4,8 rotated around the roll axis of the gripper's base, and 1,2,4 around the pitch and jaw axis). 
When using our approach with only a single orientation per test position, we note a significant decrease in required search times.
However, we can also show that the number of evaluated points is far lower, and we miss many possible grasp configurations.
Increasing the number of tested attitudes when rotating around the roll axis up to 4 also increases exploration time, but greatly increases the number of found valid poses.
Increasing to 8 no longer significantly valid poses, but still nearly doubles search times. Table \ref{table:ppo_comparison} shows how the number of sampled attitudes per position compares to PPO performance over the entire YCB dataset.
Conversely, increasing the number of tested rotations around the pitch and jaw axes does not significantly increase the number of found valid poses.
This is due to the fact that most of these grasp configurations have already been discovered by testing other relative base positions.
Due to the fact that each relative position is assigned an attitude pointing to the closest surface point, rotating around the pitch/jaw axes merely results in a grasp synthesis that targets an object surface that's already covered by a different initial relative position.

\begin{figure}
	\centering
	\begin{tabular}{|c|c|c|c|c|}
		\hline
		Att. & Quality & \# of Grasps & Misses & Time \\
		\hline \hline
		1 & 76.42\%    &  67.71\%    & 94.48\%  & 23.61\% \\
		2 & 83.91\%    &  82.48\%    & 84.68\%  & 42.81\% \\
		4 & 99.31\%    &  98.32\%    & 76.46\%  & 73.52\% \\
		8 & 100.03\%  & 100.61\%   & 75.16\% & 170.19\% \\
		\hline
	\end{tabular}
	\caption{Comparison of our approach with PPO. Rows shows the performance of our aproach when evaluating amounts of attitutes per position as a fraction of PPO performance, averaged over the YCB dataset. On average, PPO achieved a quality 0.78, with 6234.53 grasps identified, 4527.64 misses, over 34.65min per object.}
	\label{table:ppo_comparison}
\end{figure}%
	
	\section{Comparison to End-to-End Learning Methods} \label{sec:ee_comparison}%
	As mentioned in Section \ref{sec:introduction}, many recent approaches to grasp generation focus on deep learning methods that combine the entire pipeline into a single algorithm. As many components are closely interconnected, this approach has shown to be effective in rapidly generating grasps for novel objects. In the following, we briefly compare our approach as outlined in the previous sections to two end-to-end methods (Multi-fingan \cite{lundell2021multi} and Grasp'd \cite{turpin2022grasp}), showcasing advantages and disadvantages of both approaches. The first considered aspect is \textit{grasp generation time}. Here, established approaches can generate a grasp in real-time, whereas our approach requires at least a minute to find a suitable grasp for a considered object. However, once a complete pose-grasp mapping is generated, the required time is drastically reduced down to approximately 10s (most likely, this could be further reduced by improving our search algorithm). Regardless, established methods are faster. 
The next aspect is \textit{adaptability}. All approaches can handle arbitrary object shapes. However, our approach is held general and can easily be adjusted to other gripper types, regardless of whether it contains prismatic or revolute joints. End-to-end based approaches must be retrained for each new gripper type, which necessitates the generation of demonstrators or supervised learning setups, depending on the approach used. Either options requires a considerable investment of time and resources.
Additionally, we can show that our approach generates better grasps. When compared over the YCB dataset, we consistently outperform comparable methods, as shown in table \ref{table:ee_comparison}. As shown, this is consistent for various metrics, even those we were not actively monitoring during exploration.
Lastly, we considered \textit{expended gripper effort}. We compared which grasps all approaches generate and how much joint effort is required to maintain an established grasp while holding an object under normal gravity conditions. Table \ref{table:ee_torque_comp} shows the results; as can be seen, the required efforts remain comparable when considering simple shapes (e.g. balls, cubes, ...). However, when generating grasps for shapes containing ridges/ledges, requires less effort to maintain its grasp. This is due to the fact that our method takes indirect efforts into account, e.g. ledges being in contact with digit surfaces. An example is shown in Figure TODO. For this, all algorithms generated grasps on a power drill from the same initial reference pose. Our algorithm created a grasp that has the drill's motor section resting on the top surface of the gripper, whereas other approaches grasp without taking advantage of this shape feature. As a result, the end-to-end-generated grasps must compensate for gravity via exerting a sufficient amount of friction forces, whereas our grasp passively counters gravity. Similar results can be seen for most complex shapes from the YCB-dataset, leading to end-effectors requiring 4.6\% less joint efforts when employing grasps generated using our method. Note that, as mentioned above, required efforts are negligibly different when considering simple shapes without ridges.
\begin{figure*}
	\includegraphics[width=\textwidth]{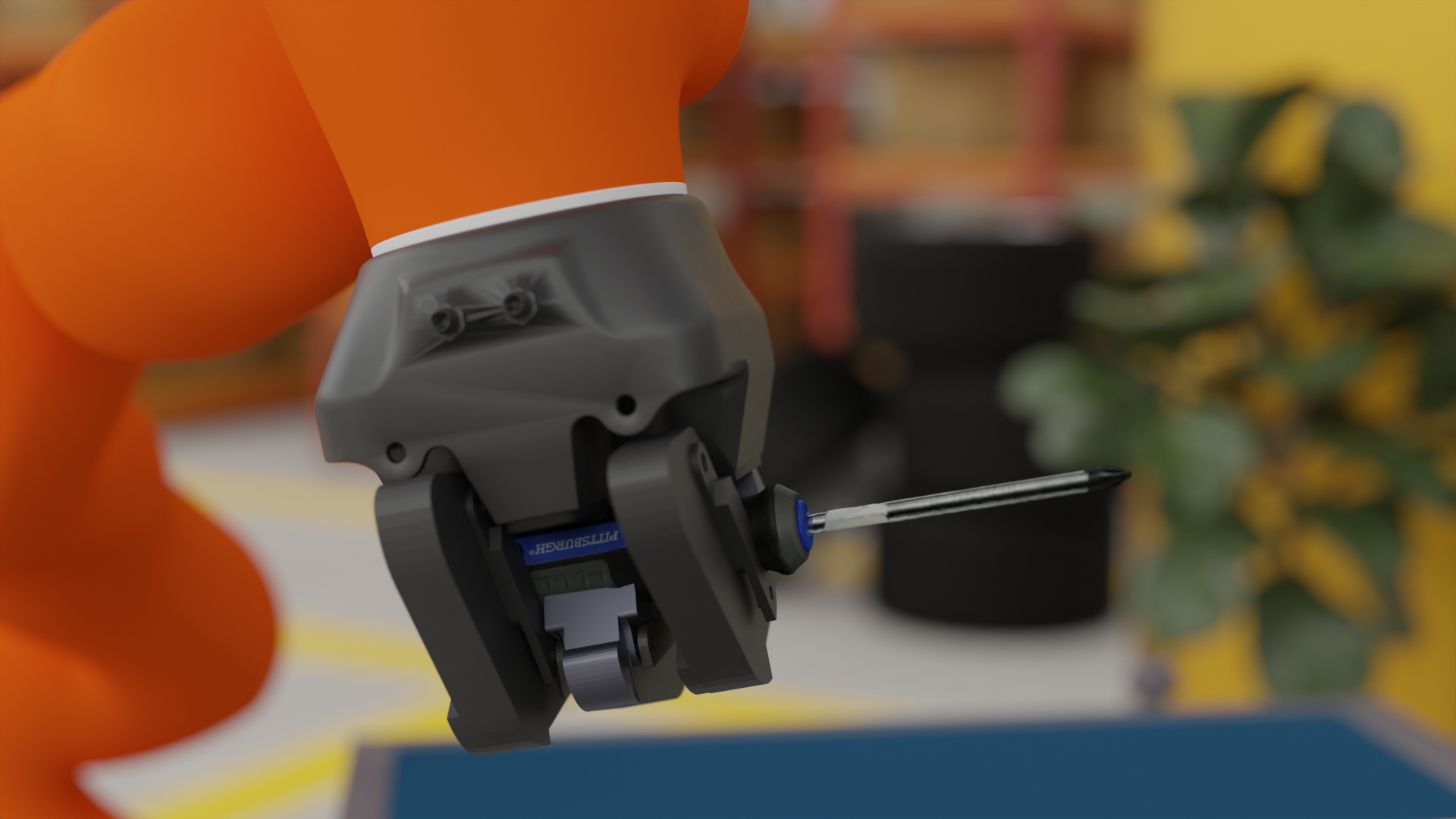}
	\caption{\texttt{Robotiq} end-effector grasping a screwdriver. Due to it's shape, only one valid grasp type is available.}
	\label{fig:money_shot}
\end{figure*}%
%
\begin{table}
	\centering
	\begin{tabular}{|c||c|c|c|}
		\hline
		Quality Metric & Our Approach & Multi-fingan & Grasp'd \\
		\hline \hline
		$\mu_{\rm s}$  & \textbf{0.86} & 0.74 & 0.75 \\
		$\mu_{\rm g}$  & \textbf{0.96} & 0.94 & 0.94 \\
		$\mu_{\rm f}$   & \textbf{0.78} & 0.73 & 0.72 \\
		$\mu_{\rm m}$ & \textbf{0.83} & 0.79 & 0.78 \\
		\hline
	\end{tabular}
	\caption{Average quality of grasps generated by end-to-end methods and our approach over the YCB dataset.}
	\label{table:ee_comparison}
\end{table}%
\begin{table}
	\centering
	\begin{tabular}{|l||c|c|c|}
		\hline
		& $\tau_{\rm s}$ & $\tau_{\rm eg}$ & $\tau_{\rm em}$\\
		\hline\hline
		\textbf{Simple Shapes} & & & \\
		\hline
		Sphere & 1.48Nm & \textbf{1.48Nm} & 1.48Nm       \\
		Cube & \textbf{1.63Nm} & 1.64Nm & \textbf{1.63Nm}         \\
		Dumbbell & \textbf{1.29Nm} & 1.53Nm & 1.29Nm     \\
		Pringles Can & \textbf{1.54Nm} & \textbf{1.54Nm} & 1.55Nm \\
		\hline
		\textbf{Small Objects} & & & \\
		\hline
		Spoon & 0.83Nm & 0.84Nm & \textbf{0.81Nm} \\
		Fork  & 0.83Nm & 0.84Nm & \textbf{0.82Nm} \\
		Screwdriver & \textbf{0.93Nm} & 0.99Nm & 1.00Nm \\
		\hline
		\textbf{Uneven Mass}   & & & \\
		\hline
		Power Drill & \textbf{1.87Nm} & 2.13Nm & 2.31Nm \\
		Hammer & 2.32Nm & 2.32Nm & \textbf{2.30Nm} \\
		\hline
		\textbf{Delicate Objects} & & & \\
		\hline
		Banana & \textbf{0.52Nm} & \textbf{0.52Nm} & 0.53Nm \\
		Tomato & \textbf{0.51Nm} & 0.53Nm & 0.53Nm \\
		Foam Ball & \textbf{0.20Nm} & \textbf{0.20Nm} & \textbf{0.20Nm} \\
		\hline
		\textbf{Textured Surface} & & & \\
		\hline
		Stanford Bunny & \textbf{0.88Nm} & 0.95Nm & 0.95Nm \\
		Spiked Ball & \textbf{1.46Nm} & \textbf{1.46Nm} & \textbf{1.46Nm} \\
		\hline
	\end{tabular}
	\caption{Required joint efforts required to maintain a generated grasp. We compare the grasps generated using our approach ($\tau_{\rm s}$), Multi-fingan ($\tau_{\rm eg}$), and Grasp'd ($\tau_{\rm em}$).}
	\label{table:ee_torque_comp}
\end{table}

	\section{Conclusion} \label{sec:conclusion}%
In this paper, we proposed a novel approach to grasp affordance learning, approaching the problem from the relative pose space. The method allows visualisation of possible grasp outcomes (both grasp types and quality) for different grasp synthesis policies. Different approaches to grasp synthesis are discussed, emphasizing the use of shape adaptation, supported by a haptic modality. Using a systematic approach to pose space exploration enables a convergence time Preliminary results show convergence times 33\% faster than those observed using PPO on identical hardware. In addition, the approach is being extended to address assembly and disassembly problems, accounting from transmission of efforts, through the grasped object, onto an assembly.


	\hypersetup{urlcolor=black}
	\bibliographystyle{IEEEtran}%
	\bibliography{IEEEabrv,references}	

\begin{thebibliography}{10}
\providecommand{\url}[1]{#1}
\csname url@samestyle\endcsname
\providecommand{\newblock}{\relax}
\providecommand{\bibinfo}[2]{#2}
\providecommand{\BIBentrySTDinterwordspacing}{\spaceskip=0pt\relax}
\providecommand{\BIBentryALTinterwordstretchfactor}{4}
\providecommand{\BIBentryALTinterwordspacing}{\spaceskip=\fontdimen2\font plus
\BIBentryALTinterwordstretchfactor\fontdimen3\font minus
  \fontdimen4\font\relax}
\providecommand{\BIBforeignlanguage}[2]{{%
\expandafter\ifx\csname l@#1\endcsname\relax
\typeout{** WARNING: IEEEtran.bst: No hyphenation pattern has been}%
\typeout{** loaded for the language `#1'. Using the pattern for}%
\typeout{** the default language instead.}%
\else
\language=\csname l@#1\endcsname
\fi
#2}}
\providecommand{\BIBdecl}{\relax}
\BIBdecl

\bibitem{golovianko2023industry}
M.~Golovianko, V.~Terziyan, V.~Branytskyi, and D.~Malyk, ``{I}ndustry 4.0 vs.
  {I}ndustry 5.0: co-existence, transition, or a hybrid,'' \emph{Procedia
  Computer Science}, vol. 217, pp. 102--113, 2023.

\bibitem{yang2023recent}
X.~Yang, Z.~Ji, J.~Wu, and Y.-K. Lai, ``Recent advances of deep robotic
  affordance learning: a reinforcement learning perspective,'' \emph{IEEE
  Transactions on Cognitive and Developmental Systems}, 2023.

\bibitem{schuh2017}
G.~Schuh, R.~Anderl, J.~Gausemeier, M.~Ten~Hompel, and W.~Wahlster,
  \emph{Industrie 4.0 Maturity Index: Managing the Digital Transformation of
  Companies}.\hskip 1em plus 0.5em minus 0.4em\relax Utz, Herbert, 2017.

\bibitem{kokic2017affordance}
M.~Kokic, J.~A. Stork, J.~A. Haustein, and D.~Kragic, ``Affordance detection
  for task-specific grasping using deep learning,'' in \emph{International
  Conference on Humanoid Robotics}.\hskip 1em plus 0.5em minus 0.4em\relax
  IEEE, 2017, pp. 91--98.

\bibitem{bicchi1995closure}
A.~Bicchi, ``On the closure properties of robotic grasping,'' \emph{The
  International Journal of Robotics Research}, vol.~14, no.~4, pp. 319--334,
  1995.

\bibitem{nguyen1988constructing}
V.-D. Nguyen, ``Constructing force-closure grasps,'' \emph{International
  Journal of Robotics Research}, vol.~7, no.~3, pp. 3--16, 1988.

\bibitem{objects1993finding}
I.-M. Chen and J.~Burdick, ``Finding antipodal point grasps on irregularly
  shaped objects,'' \emph{IEEE Transactions on Robotics and Automation},
  vol.~9, no.~4, p. 507, 1993.

\bibitem{zhu2003synthesis}
X.~Zhu and J.~Wang, ``Synthesis of force-closure grasps on 3-d objects based on
  the q distance,'' \emph{Transactions on robotics and Automation}, vol.~19,
  no.~4, pp. 669--679, 2003.

\bibitem{roa2008independent}
M.~A. Roa and R.~Su{\'a}rez, ``Independent contact regions for frictional
  grasps on 3d objects,'' in \emph{International Conference on Robotics and
  Automation}.\hskip 1em plus 0.5em minus 0.4em\relax IEEE, 2008, pp.
  1622--1627.

\bibitem{roa2009computation}
M.~A. Roa and R.~Suarez, ``Computation of independent contact regions for
  grasping 3-d objects,'' \emph{Transactions on Robotics}, 2009.

\bibitem{berenson2008grasp}
D.~Berenson and S.~S. Srinivasa, ``Grasp synthesis in cluttered environments
  for dexterous hands,'' in \emph{International Conference on Humanoid
  Robots}.\hskip 1em plus 0.5em minus 0.4em\relax IEEE, 2008, pp. 189--196.

\bibitem{abdeetedal2018grasp}
M.~Abdeetedal and M.~R. Kermani, ``Grasp synthesis for purposeful fracturing of
  object,'' \emph{Robotics and Autonomous Systems}, vol. 105, pp. 47--58, 2018.

\bibitem{feix2009comprehensive}
T.~Feix, R.~Pawlik, H.-B. Schmiedmayer, J.~Romero, and D.~Kragic, ``A
  comprehensive grasp taxonomy,'' in \emph{Robotics, science and systems:
  workshop on understanding the human hand for advancing robotic manipulation},
  vol.~2, no. 2.3, 2009, pp. 2--3.

\bibitem{zhang2018deep}
T.~Zhang, Z.~McCarthy, O.~Jow, D.~Lee, X.~Chen, K.~Goldberg, and P.~Abbeel,
  ``Deep imitation learning for complex manipulation tasks from virtual reality
  teleoperation,'' in \emph{International Conference on Robotics and
  Automation}.\hskip 1em plus 0.5em minus 0.4em\relax IEEE, 2018, pp.
  5628--5635.

\bibitem{abbeel2004apprenticeship}
P.~Abbeel and A.~Y. Ng, ``Apprenticeship learning via inverse reinforcement
  learning,'' in \emph{Proceedings of the twenty-first international conference
  on Machine learning}, 2004, p.~1.

\bibitem{xie2019learning}
X.~Xie, C.~Li, C.~Zhang, Y.~Zhu, and S.-C. Zhu, ``Learning virtual grasp with
  failed demonstrations via bayesian inverse reinforcement learning,'' in
  \emph{International Conference on Intelligent Robots and Systems}.\hskip 1em
  plus 0.5em minus 0.4em\relax IEEE, 2019, pp. 1812--1817.

\bibitem{mandi2022towards}
Z.~Mandi, F.~Liu, K.~Lee, and P.~Abbeel, ``Towards more generalizable one-shot
  visual imitation learning,'' in \emph{International Conference on Robotics
  and Automation}.\hskip 1em plus 0.5em minus 0.4em\relax IEEE, 2022, pp.
  2434--2444.

\bibitem{finn2017one}
C.~Finn, T.~Yu, T.~Zhang, P.~Abbeel, and S.~Levine, ``One-shot visual imitation
  learning via meta-learning,'' in \emph{Conference on robot learning}, 2017,
  pp. 357--368.

\bibitem{mahler2018dex}
J.~Mahler, M.~Matl, X.~Liu, A.~Li, D.~Gealy, and K.~Goldberg, ``Dex-{N}et 3.0:
  Computing robust vacuum suction grasp targets in point clouds using a new
  analytic model and deep learning,'' in \emph{International Conference on
  robotics and automation}.\hskip 1em plus 0.5em minus 0.4em\relax IEEE, 2018,
  pp. 5620--5627.

\bibitem{fang2020graspnet}
H.-S. Fang, C.~Wang, M.~Gou, and C.~Lu, ``Graspnet-1billion: A large-scale
  benchmark for general object grasping,'' in \emph{Proc. conference on
  computer vision and pattern recognition}, 2020, pp. 11\,444--11\,453.

\bibitem{morales2006anthropomorphic}
A.~Morales, P.~Azad, T.~Asfour, D.~Kraft, S.~Knoop, R.~Dillmann, A.~Kargov,
  C.~Pylatiuk, and S.~Schulz, ``An anthropomorphic grasping approach for an
  assistant humanoid robot,'' \emph{VDI BERICHTE}, vol. 1956, p. 149, 2006.

\bibitem{do2009grasp}
M.~Do, J.~Romero, H.~Kjellstr{\"o}m, P.~Azad, T.~Asfour, D.~Kragic, and
  R.~Dillmann, ``Grasp recognition and mapping on humanoid robots,'' in
  \emph{International Conference on Humanoid Robots}.\hskip 1em plus 0.5em
  minus 0.4em\relax IEEE, 2009.

\bibitem{vuong2023grasp}
A.~D. Vuong, M.~N. Vu, H.~Le, B.~Huang, B.~Huynh, T.~Vo, A.~Kugi, and
  A.~Nguyen, ``Grasp-anything: Large-scale grasp dataset from foundation
  models,'' pp. 5655--5662, 2024.

\bibitem{newbury2023deep}
R.~Newbury, M.~Gu, L.~Chumbley, A.~Mousavian, C.~Eppner, J.~Leitner, J.~Bohg,
  A.~Morales, T.~Asfour, D.~Kragic \emph{et~al.}, ``Deep learning approaches to
  grasp synthesis: A review,'' \emph{IEEE Transactions on Robotics}, 2023.

\bibitem{liu1999qualitative}
Y.-H. Liu, ``Qualitative test and force optimization of 3-d frictional
  form-closure grasps using linear programming,'' \emph{IEEE Transactions on
  Robotics and Automation}, vol.~15, no.~1, pp. 163--173, 1999.

\bibitem{lu2023closer}
W.~Lu, X.~Zhao, S.~Magg, M.~Gromniak, M.~Li, and S.~Wermterl, ``A closer look
  at reward decomposition for high-level robotic explanations,'' in \emph{IEEE
  International Conference on Development and Learning}, 2023, pp. 429--436.

\bibitem{mordatch2012discovery}
I.~Mordatch, E.~Todorov, and Z.~Popovi{\'c}, ``Discovery of complex behaviors
  through contact-invariant optimization,'' \emph{ACM Transactions on Graphics
  (ToG)}, vol.~31, pp. 1--8, 2012.

\bibitem{mordatch2012contact}
I.~Mordatch, Z.~Popovi{\'c}, and E.~Todorov, ``Contact-invariant optimization
  for hand manipulation,'' in \emph{Proceedings of the ACM
  SIGGRAPH/Eurographics symposium on computer animation}, 2012, pp. 137--144.

\bibitem{turpin2022grasp}
D.~Turpin, L.~Wang, E.~Heiden, Y.-C. Chen, M.~Macklin, S.~Tsogkas,
  S.~Dickinson, and A.~Garg, ``Grasp’d: Differentiable contact-rich grasp
  synthesis for multi-fingered hands,'' in \emph{European Conference on
  Computer Vision}.\hskip 1em plus 0.5em minus 0.4em\relax Springer, 2022, pp.
  201--221.

\bibitem{lundell2021multi}
J.~Lundell, E.~Corona, T.~N. Le, F.~Verdoja, P.~Weinzaepfel, G.~Rogez,
  F.~Moreno-Noguer, and V.~Kyrki, ``Multi-fingan: Generative coarse-to-fine
  sampling of multi-finger grasps,'' in \emph{International Conference on
  Robotics and Automation}.\hskip 1em plus 0.5em minus 0.4em\relax IEEE, 2021,
  pp. 4495--4501.

\bibitem{zechmair2021assessing}
M.~Zechair and Y.~Morel, ``Assessing grasp quality using local sensitivity
  analysis,'' in \emph{Int. Conf. on Intelligent Robots and Systems}.\hskip 1em
  plus 0.5em minus 0.4em\relax IEEE, 2021.

\bibitem{zechmair2021penalty}
M.~Zechmair and Y.~Morel, ``Penalty-based numerical representation of rigid
  body interactions with applications to simulation of robotic grasping,'' in
  \emph{International Conference on Electrical, Computer, Communications and
  Mechatronics Engineering}.\hskip 1em plus 0.5em minus 0.4em\relax IEEE, 2022,
  pp. 1--8.

\bibitem{calli2017yale}
B.~Calli, A.~Singh, J.~Bruce, A.~Walsman, K.~Konolige, S.~Srinivasa, P.~Abbeel,
  and A.~M. Dollar, ``Yale-cmu-berkeley dataset for robotic manipulation
  research,'' \emph{International Journal of Robotics Research}, vol.~36,
  no.~3, pp. 261--268, 2017.

\bibitem{roa2015grasp}
M.~A. Roa and R.~Su{\'a}rez, ``Grasp quality measures: review and
  performance,'' \emph{Autonomous robots}, vol.~38, no.~1, pp. 65--88, 2015.

\bibitem{morio2011global}
J.~Morio, ``Global and local sensitivity analysis methods for a physical
  system,'' \emph{European journal of physics}, vol.~32, no.~6, p. 1577, 2011.

\bibitem{wang2020truly}
Y.~Wang, H.~He, and X.~Tan, ``Truly proximal policy optimization,'' in
  \emph{Uncertainty in Artificial Intelligence}, 2020, pp. 113--122.

\bibitem{li1988task}
Z.~Li and S.~S. Sastry, ``Task-oriented optimal grasping by multifingered robot
  hands,'' \emph{IEEE Journal on Robotics and Automation}, vol.~4, no.~1, pp.
  32--44, 1988.

\bibitem{kim2001optimal}
B.-H. Kim, S.-R. Oh, B.-J. Yi, and I.~H. Suh, ``Optimal grasping based on
  non-dimensionalized performance indices,'' in \emph{Proceedings 2001 IEEE/RSJ
  International Conference on Intelligent Robots and Systems. Expanding the
  Societal Role of Robotics in the the Next Millennium (Cat. No. 01CH37180)},
  vol.~2.\hskip 1em plus 0.5em minus 0.4em\relax IEEE, 2001, pp. 949--956.

\bibitem{gualtieri2016high}
M.~Gualtieri, A.~Ten~Pas, K.~Saenko, and R.~Platt, ``High precision grasp pose
  detection in dense clutter,'' in \emph{International Conference on
  Intelligent Robots and Systems}.\hskip 1em plus 0.5em minus 0.4em\relax IEEE,
  2016, pp. 598--605.

\bibitem{ten2017grasp}
A.~Ten~Pas, M.~Gualtieri, K.~Saenko, and R.~Platt, ``Grasp pose detection in
  point clouds,'' \emph{International Journal of Robotics Research}, vol.~36,
  no. 13-14, pp. 1455--1473, 2017.

\bibitem{liang2019pointnetgpd}
H.~Liang, X.~Ma, S.~Li, M.~G{\"o}rner, S.~Tang, B.~Fang, F.~Sun, and J.~Zhang,
  ``Pointnetgpd: Detecting grasp configurations from point sets,'' in
  \emph{International Conference on Robotics and Automation}.\hskip 1em plus
  0.5em minus 0.4em\relax IEEE, 2019, pp. 3629--3635.

\bibitem{aleotti20123d}
J.~Aleotti and S.~Caselli, ``A 3d shape segmentation approach for robot
  grasping by parts,'' \emph{Robotics and Autonomous Systems}, vol.~60, no.~3,
  pp. 358--366, 2012.

\bibitem{zechmair2023active}
M.~Zechmair and Y.~Morel, ``Active electric perception-based haptic modality
  with applications to robotics,'' in \emph{International Conference on
  Intelligent Robots and Systems}.\hskip 1em plus 0.5em minus 0.4em\relax IEEE,
  2023, pp. 5950--5957.

\bibitem{fabri2009cgal}
A.~Fabri and S.~Pion, ``{CGAL}: The computational geometry algorithms
  library,'' in \emph{Proceedings of the 17th ACM SIGSPATIAL international
  conference on advances in geographic information systems}, 2009, pp.
  538--539.

\bibitem{schulman2017proximal}
J.~Schulman, F.~Wolski, P.~Dhariwal, A.~Radford, and O.~Klimov, ``Proximal
  policy optimization algorithms,'' \emph{arXiv preprint arXiv:1707.06347},
  2017.

\bibitem{fan2018real}
Y.~Fan, T.~Tang, H.-C. Lin, and M.~Tomizuka, ``Real-time grasp planning for
  multi-fingered hands by finger splitting,'' in \emph{2018 IEEE/RSJ
  International Conference on Intelligent Robots and Systems (IROS)}.\hskip 1em
  plus 0.5em minus 0.4em\relax IEEE, 2018, pp. 4045--4052.

\bibitem{zhang2022simulation}
Z.~Zhang and C.~Zheng, ``Simulation of robotic arm grasping control based on
  proximal policy optimization algorithm,'' in \emph{Journal of Physics:
  Conference Series}, vol. 2203, no.~1.\hskip 1em plus 0.5em minus 0.4em\relax
  IOP Publishing, 2022, p. 012065.

\bibitem{pedersen2020grasping}
O.-M. Pedersen, E.~Misimi, and F.~Chaumette, ``Grasping unknown objects by
  coupling deep reinforcement learning, generative adversarial networks, and
  visual servoing,'' in \emph{2020 IEEE international conference on robotics
  and automation (ICRA)}.\hskip 1em plus 0.5em minus 0.4em\relax IEEE, 2020,
  pp. 5655--5662.

\bibitem{zheng2023object}
Q.~Zheng, Z.~Peng, P.~Zhu, Y.~Zhao, R.~Zhai, and W.~Ma, ``An object recognition
  grasping approach using proximal policy optimization with yolov5,''
  \emph{IEEE Access}, 2023.

\bibitem{weidler2023angorapy}
\BIBentryALTinterwordspacing
T.~Weidler and M.~Senden, ``{AngoraPy - Anthropomorphic Goal-Oriented Robotic
  Control for Neuroscientific Modeling},'' 2023. [Online]. Available:
  \url{https://doi.org/10.5281/zenodo.6636482}
\BIBentrySTDinterwordspacing

\end{thebibliography}
\end{document}